\documentclass[aps,amsmath,amssymb,reprint,superscriptaddress]{revtex4-1}
\usepackage{dcolumn}% Align table columns on decimal point
\usepackage{bm}% bold math
\usepackage{amsmath}
\usepackage{amssymb}
\usepackage{amsfonts}
\usepackage{subfigure}
\usepackage[pdftex]{graphicx}
\usepackage{booktabs}
\usepackage[top=0.75in, bottom=0.75in, left=0.75in, right=0.75in]{geometry}
\begin{document}

\title{\textbf{Inferring the time-varying coupling of dynamical systems with temporal convolutional autoencoders}}

\author{Josuan Calderon}
\affiliation{Department of Physics, Emory University, Atlanta, GA, 30322}
\author{Gordon J. Berman}\email{gordon.berman@emory.edu}
\affiliation{Department of Physics, Emory University, Atlanta, GA, 30322}
\affiliation{Department of Biology, Emory University, Atlanta, GA, 30322}
\date{ \today} 

\begin{abstract}
Most approaches for assessing causality in complex dynamical systems fail when the interactions between variables are inherently non-linear and non-stationary. Here we introduce Temporal Autoencoders for Causal Inference (TACI), a methodology that combines a new surrogate data metric for assessing causal interactions with a novel two-headed machine learning architecture to identify and measure the direction and strength of time-varying causal interactions. Through tests on both synthetic and real-world datasets, we demonstrate TACI's ability to accurately quantify dynamic causal interactions across a variety of systems. Our findings display the method's effectiveness compared to existing approaches and also highlight our approach's potential to build a deeper understanding of the mechanisms that underlie time-varying interactions in physical and biological systems.
\end{abstract}

\maketitle

\section*{Introduction}
\label{sec:introduction}

Rather than being static in time, interactions between parts of a complex system continuously ebb and flow, with one variable driving another at one point, just to see the relationship reverse or lessen or disappear at a later point in time.  Real-world signals are seldom stationary and well-behaved, and their causal linkages and interactions frequently appear, disappear, and reappear, possibly changing in strength over time. Examples of such systems abound in neuroscience \cite{Time_Varying_causality_EEG_data,buzsaki2006rhythms}, ecology \cite{sardine_anchovies_competition}, finance, \cite{Time_varying_causality_between_crude_oil_and_stock_markets, Time_varying_causality_between_renewable_and_non_renewable_energy, Time_Varying_causality_green_bonds_financial}, and climate \cite{ENSO_Greenhouse, ENSO_Decadal_variability,Synchronization_and_causality_Sugihara}.

Despite the ubiquity of these dynamically altering interactions, however, most methods for assessing causality in complex dynamical systems have difficulty measuring how the direction and extent of interactions between variables in a system alter in time.  This difficulty arises from several inherent characteristics of complex systems and the limitations of existing causal assessment methodologies. Most of these methods, including Granger Causality (GC) \cite{Granger_original}, often assume that the dynamical system should be approximately stationary, meaning their statistical properties do not change over time. Other common assumptions, such as linearity and time invariance are also often violated in real-world complex systems. These constraints significantly limit the applicability and accuracy of these approaches in many scenarios \cite{Runge_PCMCI, tcdf_attention, Granger_problems, Granger_temporal_problems_econ, causality_biological_time_series, Sugihara_Detecting_Causality_in_Complex_Ecosystems}. 

In addition, in systems where variables are strongly coupled and synchronized, some of these causality inference methods struggle to accurately infer the coupling strength and direction of causality \cite{Sugihara_Detecting_Causality_in_Complex_Ecosystems}. This issue extends to scenarios of intermediate coupling, where the variables are neither weakly nor strongly linked. Additionally, the presence of noise in the system leads to a decrease in cross-mapping fidelity, revealing further limitations \cite{ccm_fails, Sugihara_time_delayed_interactions_ccm}. Although the lack of correlation is neither necessary nor sufficient to demonstrate causation \cite{Sugihara_corr_no_causation_1, Sugihara_corr_no_causation_2}, correlation does play an important role in many statistical methods as the basis for hypothesis tests for causality. Mirage correlations can appear in the simplest nonlinear systems \cite{mysterud2001nonlinear}. Variables that may be positively correlated at some point in time can become anti-correlated some moments after or even lose all coherence. However, most causality methods do not adequately account for the fact that sudden changes in correlation over time between variables may indicate a change in the underlying temporal causal relationships. 

In an attempt to overcome some of these limitations, here we introduce a new methodology for probing time-varying causal interactions using a new metric for assessing causal interactions combined with a novel machine learning architecture for causal inference, which we call Temporal Autoencoders for Causal Inference (TACI).  We show the method's effectiveness on synthetic and real-world data sets, both in an absolute sense and in comparison to extant methods, particularly focused on how to find time-varying causal structure in complex dynamical systems.

\section*{Overview of Methodology}
In our methodology, we adopt a two-fold approach towards developing a causal inference method that accurately assesses causality between variables $x(t)$ and $y(t)$ in the Granger sense for nonlinear systems with time-varying interactions.  The first aspect of our approach is to use a novel surrogate data comparison metric -- the Comparative Surrogate Granger Index (CSGI) -- that measures the relative improvement in prediction accuracy when including both variables vs. one of them and a randomized version of the other. The other aspect is to use a two-headed Temporal Convolutional Network architecture to robustly capture the space of potential nonlinear mappings between variables across the entire time series. As will be observed, the CSGI with linear autoregressive models works well to identify causal interactions in situations where relatively straightforward mappings exist between variables, but the more complicated neural network model is more effective when the mappings are more non-linear.

\subsection*{Comparative Surrogate Granger Index (CSGI)}
Informally, Granger Causality (GC) defines a causal interaction from $x(t)$ to $y(t)$ to be when knowing the full history of both $x(t)$ and $y(t)$ provides a better prediction about the future of $y(t)$ than knowing just the history of $y(t)$ alone.  While there are many variations on this methodology \cite{extended_granger_chen, Kernel_granger, Tank_Neural_Granger}, the typical form used is to compare two models of similar type (e.g., linear autoregressive models, feedforward neural networks, etc.) based on their ability to predict the future state of $y(t)$.  More explicitly, the comparison is between 
\begin{equation}\label{f_eqn}
y(t) \approx f(y_{t-1},y_{t-2},\ldots)
 \end{equation} 
 and 
 \begin{equation}\label{g_eqn}
 y(t) \approx g(x_{t-1},x_{t-2},\ldots,y_{t-1},y_{t-2},\ldots).  
 \end{equation}
 Usually, an F-test is used to determine whether the latter model is preferred over the former.  

This form, however, suffers from two limitations.  First, the comparison is a binary one -- the second model is ``significantly" better than the first or it is not -- thus, differences in the strength of coupling can not be detected, just the presence or absence. Second, because the F-test and similar methods incorporate strong assumptions about the underlying dynamics of the system, statistical statements deriving from these tests are often not robust under resampling or re-parameterization.  In addition, because the model complexities for the two models being compared are inevitably quite different, with one typically having twice as many parameters as the other, the F-test often fails to detect causal interactions properly.  

A common strategy for ameliorating these limitations is to compare not  
\begin{flalign} \nonumber
    \hspace{0.5in}&\mbox{$f(y_{t-1},y_{t-2},\ldots)$} &&\\  \nonumber
    & \hspace{0.2in}\mbox{and} &&\\ \nonumber
    &\mbox{$g(x_{t-1},x_{t-2},\ldots,y_{t-1},y_{t-2},\ldots)$}, 
\end{flalign}
but rather
\begin{flalign}\label{model_eqn}\nonumber
    \hspace{0.5in}&\mbox{$f(x_{t-1},x_{t-2},\ldots,y_{t-1},y_{t-2},\ldots)$} &&\\ \nonumber
    & \hspace{0.2in}\mbox{and} &&\\ \nonumber
    &\mbox{$g(x^{(s)}_{t-1},x^{(s)}_{t-2},\ldots,y_{t-1},y_{t-2},\ldots)$},
\end{flalign}
where $x^{(s)}$ is a \emph{surrogate} data set that shares similar statistical properties to $x(t)$ but is shuffled in some manner (e.g., shuffling values to preserve the distribution of values or shuffling phases of the time series' Fourier Transform to preserve the frequency profile), or even a completely randomized time series.  Typically, this comparison is accomplished through the Extended Granger Causality Index (EGCI) \cite{extended_granger_chen, Extended_Granger_Schiatti_2015}.  If $\epsilon_y(t)$ are the residuals for fitting the future of $y(t)$ on the past of $y(t)$ and $\epsilon_{xy}(t)$ are the residuals for fitting the future of $y(t)$ on the pasts of both $x(t)$ and $y(t)$, then the EGCI is given by ratio between the relative reduction in residual variance when the past of $x(t)$ is included in the model:
\begin{equation}
\mbox{EGCI} = 1 - \frac{\mbox{var}(\epsilon_{xy}(t))}{\mbox{var}(\epsilon_{y}(t))}.
\end{equation}
$y(t)$ is thus said to cause $x(t)$ if the EGCI using the actual values of $x(t)$ is significantly higher than the EGCI found substituting $x(t)$ for $x^{(s)}(t)$.  

Our approach attempts to assess directly the relative increase in variance explained for the predictive model when using $x(t)$ vs. $x^{(s)}(t)$.  Specifically, if $R^2_{xy}$ is the fraction of variance explained about the future of $y(t)$ using the pasts of $x(t)$ and $y(t)$ in the model and $R^2_{x^{(s)} y}$ is the fraction of variance explained using $x^{(s)}(t)$ and $y(t)$, then we define the Comparative Surrogate Granger Index (CSGI), $\chi_{x\to y}$, to be defined via
\begin{equation}\label{CSGI}
\chi_{x\to y} = \frac{R^2_{xy}-R^2_{x^{(s)} y}}{\frac{1}{2}(R^2_{xy}+R^2_{x^{(s)} y})}.
\end{equation}
This metric's advantages over the EGCI are that it is able to measure small changes in causal interactions and that it explicitly measures the difference in predictive power between using actual data and using surrogate data to predict the future.  In this article, we will be measuring $\chi_{x\to y}$ and $\chi_{y\to x}$ for all pairs of variables to assess whether there is causal coupling between two variables, whether it is uni- or bi-directional, and the relative strength of the coupling.

\subsection*{Temporal Autoencoders for Causal Inference}
While one advance in our methodology is the use of the CSGI in the previous section, the other novel contribution is the use of a new artificial neural network architecture to calculate the functions $f$ and $g$ that are used to predict the future of $y(t)$. The original (and still most common) models (\cite{Granger_original}) for $f$ and $g$ are auto-regressive linear models of the form
\begin{equation}\label{linear_granger}
    y(t) = \sum_{i=1}^k (a_i x(t-i) + b_i y(t-i)) + \epsilon_t,
\end{equation}
where $x^{(s)}(t)$ can be substituted for $x(t)$ when using the surrogate approach. While this relatively simple approach shows impressive performance in a variety of scenarios, these models fail to accurately predict known causal interactions for couplings that have weak to moderate coupling and are governed by nonlinear dynamics that are not well approximated by linear models \cite{Granger_1969, Pearl_book}. This inability is typically because the systems fail to satisfy separability.  In other words, all information about a causative factor has to be inherent to that specific variable and can be omitted by removing that variable from the model, as is the case for purely stochastic or linear systems \cite{Sugihara_Detecting_Causality_in_Complex_Ecosystems}. For systems with strongly nonlinear deterministic components, however, this assumption fails, and, accordingly, so do the predictions from auto-regressive linear GC \cite{Granger_problems,Pearl_book}. In addition, because these linear models have difficulty predicting information across multiple timescales, they often have difficulty detecting subtle shifts in causality as a function of time.

In recent years, a solution has been to replace the linear model in (\ref{linear_granger}) with deep neural networks of varying architectures that, due to their expressive nature, excel in approximating complex functions \cite{Goodfellow_deep_learning}. These methods include the use of Variational Autoencoders to estimate causal effects \cite{VAE_causal}, Causal Generative Neural Networks to learn functional causal models \cite{GAN_causal}, Neural Granger to estimate non-linearly dependencies based on Granger causality principles \cite{Tank_Neural_Granger}, and the Temporal Causal Discovery Framework (TCDF) to address time delay causal relationships \cite{tcn_attention}. These methods, however, are often unwieldy to train, are prone to overfitting, and are susceptible to inaccuracies in the presence of a significant amount of noise.  

\begin{figure*}
  \centering
  \includegraphics[width=13cm]{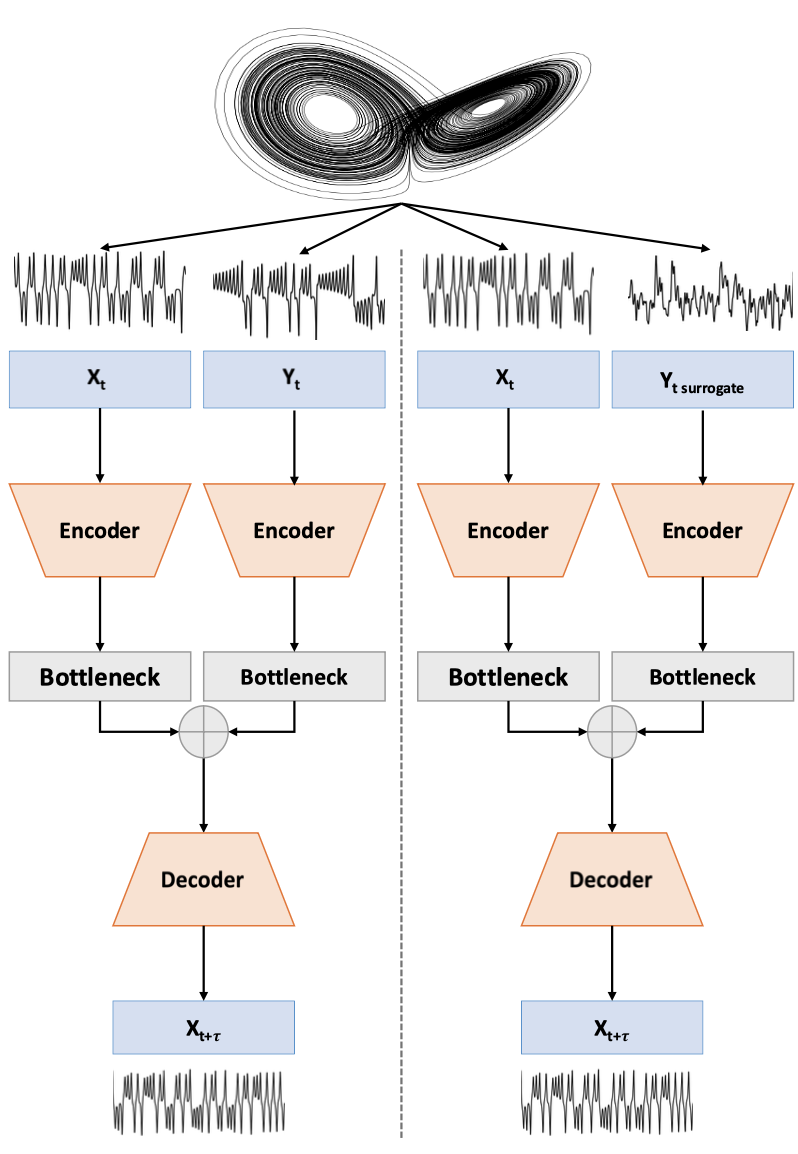}
  \caption[Schematic of the Temporal Autoencoders for Causal Inference (TACI) Networks]{Schematic of the Temporal Autoencoders for Causal Inference (TACI) Networks.  We use a two-headed network consisting of Temporal Convolutional Networks that interact through a shared latent space to predict a time-shifted version of one of the two input time series.  For each pair of variables we wish to examine (here, $X$ and $Y$), we train two networks for each causal direction: one using $X$ and $Y$ as inputs and another using $X$ and a randomized version of $Y$. We consider an interaction from $Y\to X$ to be causal if the network using the actual value of $Y$ predicts the future of $X$ better than the network using the surrogate version of $Y$. In this particular case, we show the approach applied to two different variables from the Lorenz system.}
  \label{fig:taci_autoencoder_model}
\end{figure*}

In this paper, we introduce a novel neural network architecture for causality using a two-headed Temporal Convolutional Network (TCN) autoencoder (Fig. \ref{fig:taci_autoencoder_model}).  TCNs, the primary building block of our approach, are a specialized type of neural network that integrates causal inference into convolutional architectures. First introduced for video-based action segmentation \cite{LeaColin_TCN}, TCNs quickly became popular due to their ability to extend most of the convolutional advantages of regular CNNs -- including sparsity and translational equivariance -- into the time domain through a series of dilated and causal convolutional layers.  A key characteristic of this framework is its simplicity, relatively long memory, and ability to outperform most convolutional architectures in auto-regressive prediction tasks \cite{evaluation_TCN}. 

Our approach, which we call Temporal Autoencoders for Causal Inference (TACI) is a neural network that consists of a two-headed TCN autoencoder, where two TCNs are used to encode time series $x(t)$ and $y(t)$, and a third is used for decoding an equivalently long time series describing the future trajectory of $y(t)$ (shifted by some time, $\tau$) from a relatively low-dimensional latent space that is derived from the outputs of the first two autoencoders.  A more detailed description of our model and our training methodology can be found in \emph{Materials and methods}.  Code is available here: \href{https://github.com/bermanlabemory/Temporal-Autoencoders-For-Causal-Inference-TACI}{https://github.com/bermanlabemory/Temporal-Autoencoders-For-Causal-Inference-TACI}.

For each comparison of interest, we train four versions of this network: one using $x(t)$ and $y(t)$ as input time series to predict the future of $x(t)$, another that is the same except for replacing $x(t)$ with the surrogate data $x^{(s)}(t)$, another pair of the networks that are structured the same except with $x$ and $y$ reversed in each case.  Given these four trained networks, we can then make predictions for the future of the appropriate variable and calculate the fraction of variance explained over a moving window (i.e., $R^2_{xy}(t)$ or $R^2_{x^{(s)} y}(t)$).  From these values, we can then apply Eqn. (\ref{CSGI}) to calculate the CSGI values 
$\chi_{x\to y}$ and $\chi_{y\to x}$, which we will use to assess causal inference between these two variables.

\subsection*{Other Methods We Compare Against}
Alongside comparing GC with linear autoregressive models, which we will refer to here as Surrogate Linear Granger Causality (SLGC), we will also test against two other commonly used methods: Convergent Cross Mapping and Transfer Entropy.

\subsubsection*{Convergent Cross Mapping (CCM)}
Convergent Cross Mapping (CCM) was introduced to determine causation in systems that could be modeled as relatively noiseless deterministic dynamical systems \cite{Sugihara_Detecting_Causality_in_Complex_Ecosystems}. The core concept of this approach is that according to Takens' Embedding Theorem, if  $x(t)$ and $y(t)$ are two variables of a deterministic dynamical system, one can reconstruct $x(t)$ from a delay embedding of $y(t)$ if and only if the time derivative of $x(t)$ explicitly depends on $y(t)$ \cite{Sugihara_manifolds, Takens_original, Sugihara_nonlinear_state_space_reconstruction}. Thus, if it is possible to predict $x(t)$ from an embedding of $y(t)$ alone, we would say that $y$ has a causal interaction with $x$. Practically, these predictions are calculated by predicting $x(t)$ from $y(t)$ (and vice versa) and computing the correlation coefficient between the actual and predicted values \cite{Sugihara_Detecting_Causality_in_Complex_Ecosystems}, with correlations near one implying a strong casual influence and correlations near zero implying no or little influence. Here, we used Scikit Convergent Cross Mapping (skccm), a Python-based-library implementation of CCM for causal discovery \cite{skccm}. 

\subsubsection*{Transfer Entropy}
Transfer entropy (TE) is a metric that quantifies a reduction in uncertainty in predicting the future of one variable given the past of another using formalism from information theory \cite{transfer_entropy_Schreiber}. Specifically, we can measure the transfer entropy from $y(t)$ to $x(t)$ at a given distance in the future, $\tau$, ($T_{Y\to X}(\tau)$) via 
\begin{align}
    T_{Y\to X}(\tau) &= H(x(t)\vert x(t-1),\ldots, x(t-\tau)) \nonumber \\ 
    &- H(x(t)\vert  x(t-1),\ldots, x(t-\tau), \\ \nonumber
    & \hspace{1in} y(t-1),\ldots,y(t-\tau)), 
\end{align}
where $H(X\vert Y)$ is the Shannon entropy of the conditional probability distribution $p(X|Y)$.  This quantity is zero if adding information about the past of $y(t)$ results in no reduction in our future guesses for $x(t)$, and if it is non-zero, the quantity can be interpreted as the rate of information flowing from $Y$ to $X$.  Practically, we calculate TE for our systems using the Java Information Dynamics Toolkit (JIDT or Infodynamics Toolkit) \cite{JIDT_transfer_entropy}.

\section*{Results}

\subsection*{Artificial Test Systems}
To test the validity of our approach, we applied the methodology to a variety of different deterministic and stochastic dynamical system models with known causal interactions, finding that TACI performs well across all cases. In particular, we are interested in cases where the coupling changes in time, which we will explore in detail for the Coupled Henon Maps system.

\subsubsection*{The R\"{o}ssler-Lorenz System}
\begin{figure*}
  \centering
  \includegraphics[width=13cm]{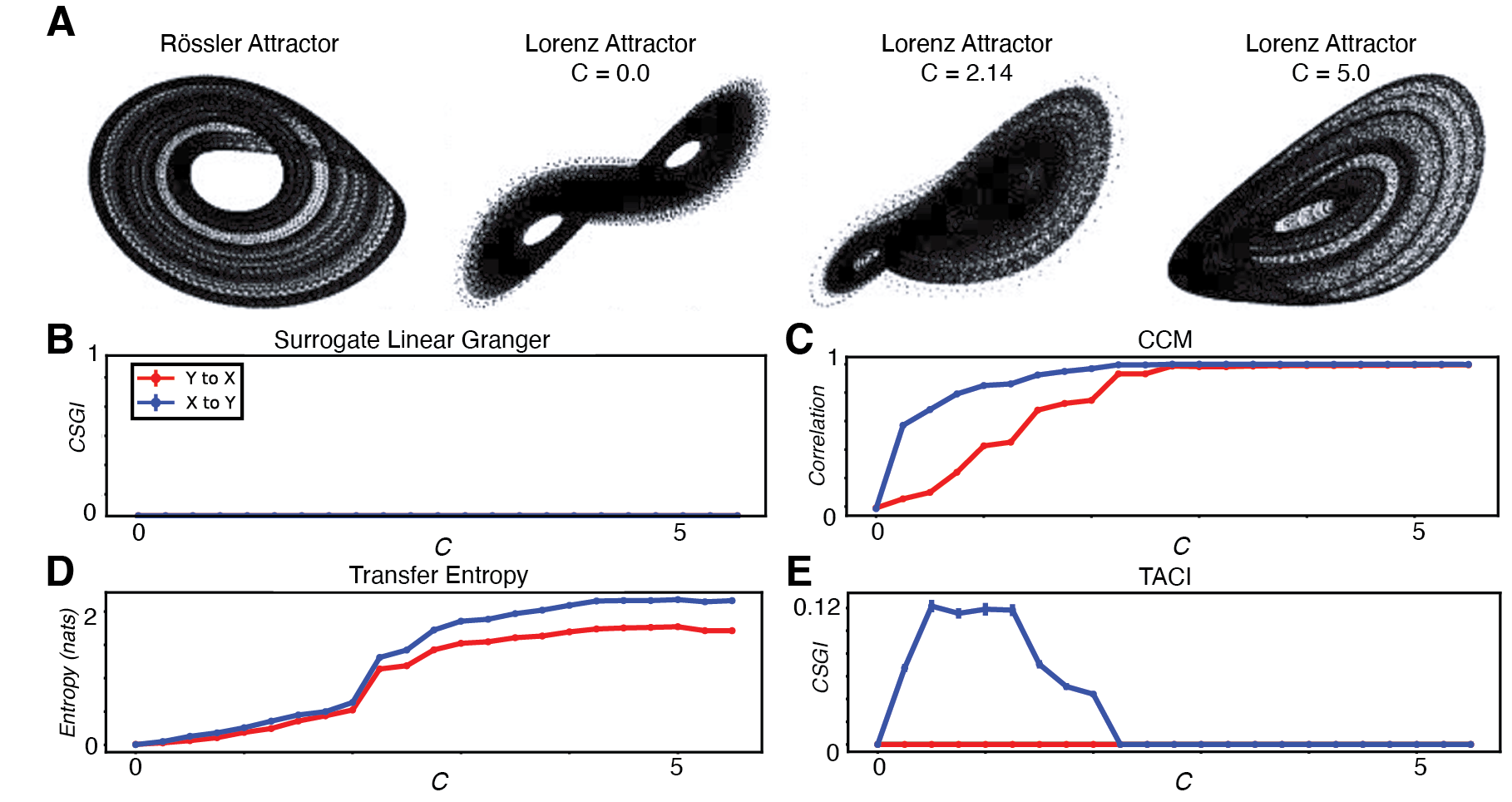}
  \caption[Causal inference in the R\"{o}ssler-Lorenz System]{Causal inference in the R\"{o}ssler-Lorenz System. \textbf{A}) 2-dimensional projections of the R\"{o}ssler attractor (left) and the Lorenz system (right three plots) as $C$ increases.  Mathematically, there is only coupling from $X\to Y$, but starting near $C=2.14$, the two systems become synchronized, making finding the causal interactions an ill-posed problem. \textbf{B-E}) Results from applying the four methods to the system.  Note that only TACI accurately predicts the unidirectional coupling in the regime above $C>0$ and before synchronization occurs. Error bars are generated using a bootstrapping procedure (see Materials and Methods).}
  \label{fig:rossler-lorenz}
\end{figure*}

Our first example case is a system of coupled chaotic attractors, where the Lorenz system ($\vec{y}$) \cite{Lorenz_original} is driven by a R\"ossler oscillator ($\vec{x}$)\cite{Rossler_original}:
\begin{align}
    \dot{x}_1 &= -6(x_2 + x_3), \nonumber \\ \nonumber
    \dot{x}_2 &= 6(x_1 + 0.2x_2), \\ \nonumber
    \dot{x}_3 &= 6\Big(0.2 + x_3(x_1 - 5.7)\Big), \\ \nonumber
    \dot{y}_1 &= 10(y_2 - y_1), \\
    \dot{y}_2 &= 28y_1 - y_2 - y_1y_3 + C x^2_2, \\ \nonumber
    \dot{y}_3 &= y_1 y_2 - \frac{8}{3} y_3 \nonumber,
\end{align}
where the constant $C$ controls the coupling strength of the system, and the driving severely distorts the behavior of the Lorenz attractor as $C$ increases \cite{rossler_lorenz_Le_Van_Quyen}. Synchronization between $\vec{x}$ and $\vec{y}$ starts near $C=2.14$, making the two systems' behavior effectively coupled above this point despite the lack of an explicit coupling term, making traditional formal notions of causality ill-posed (Fig. \ref{fig:rossler-lorenz}A) \cite{Learning_driver_response_relationships_from_synchronization}. The solutions to the differential equations were generated by using a fourth-order Runge-Kutta method. $C$ was chosen between 0 and 5, computing a time series of length 300,000 ($dt = 0.1$) after a burn-in time of 30,000 time points at each coupling strength.

As seen in Figure \ref{fig:rossler-lorenz}B-E, TACI is the only method of the four tried here that accurately predicts the unidirectional coupling from $\vec{x}$ to $\vec{y}$.   SLGC (Fig. \ref{fig:rossler-lorenz}B) fails to predict any coupling whatsoever between the variables, and CCM and TE (Figs. \ref{fig:rossler-lorenz}C-D) predict bidirectional coupling (albeit with somewhat more information flowing from $\vec{x}\to\vec{y}$ than in the reverse). TACI, in contrast, predicts only unidirectional coupling until the point of synchronization ($C\approx 2.14$), after which, it predicts no effective causation in either direction.

\subsubsection*{Coupled Bi-directional Two-Species Model}
In contrast to the R\"{o}ssler-Lorenz System, the bi-directional two-species model \cite{Henon_original}, is calculated in discrete time, and it exhibits (unsurprisingly) bi-directional coupling: 
\begin{align}
    x(t+1) &= x(t)\Big( 3.8 - 3.8x(t) - Cy(t)\Big) \nonumber \\
    y(t+1) &= y(t)\Big( 3.5 - 3.5y(t) - 5Cx(t)\Big).
\end{align}
where $C$ once again is the coupling strength, noting that the coupling strength is five times larger from $x\to y$ than in the reverse direction. In this system, separability is not satisfied (i.e., information about $y$  is redundantly present in $x$ and vice versa). Despite the fact this model is deterministic and dynamically coupled, it shows alternating periods of positive, negative, and zero correlation \cite{Sugihara_Detecting_Causality_in_Complex_Ecosystems}. For values of $C\in[0,0.35]$, we created a bivariate time series of length 300,000 (after a burn-in of 30,000 time points). The initial conditions were generated with random starting points drawn from the uniform distribution $(0.01, 0.99)$.

\begin{figure*} 
  \centering
  \includegraphics[width=13cm]{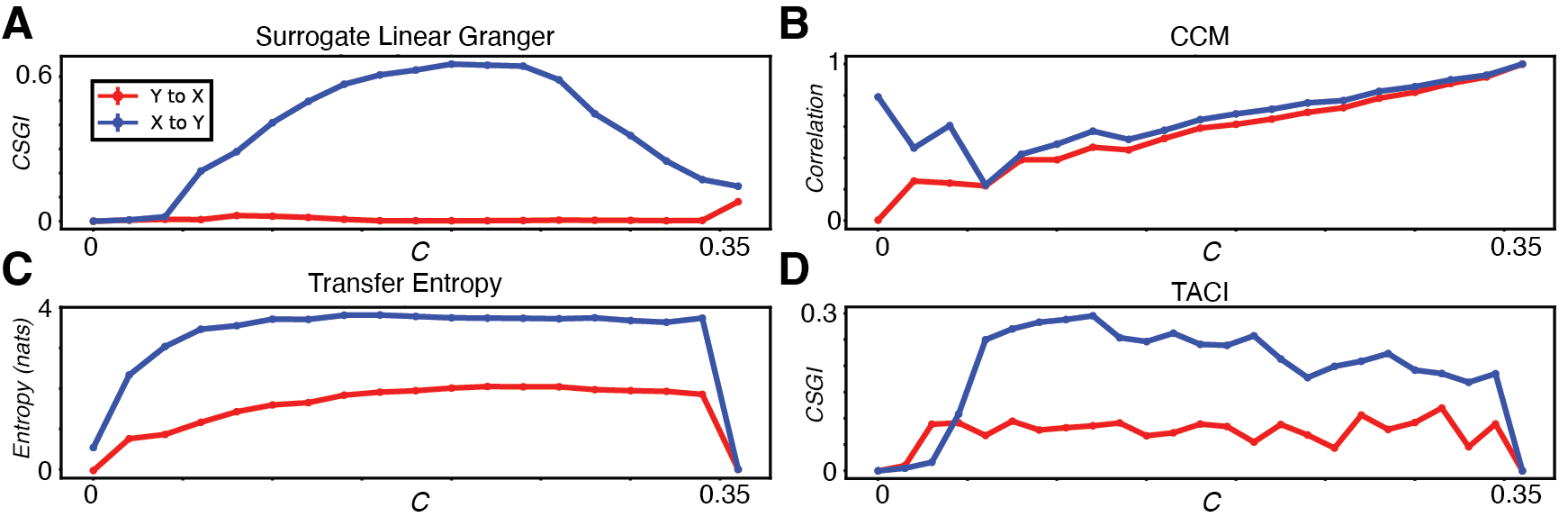}
  \caption[Causal inference in the bidirectional species system]{Causal inference in the bidirectional species system. \textbf{A-D}) Results from applying the four methods to the bidirectional species system.  Error bars are generated using a bootstrapping procedure (see Materials and Methods).}
  \label{fig:bidirectional-species}
\end{figure*}
Applying the four methods to these data (Fig. \ref{fig:bidirectional-species}), we find that both TE and TACI correctly identify both the bi-directional aspect of the coupling and the increased causal link from $x\to y$ compared to $y\to x$.  CCM identifies the bi-directionality correctly, but it does not identify the relative strength of the couplings, and SLGC is unable to identify any causal link from $y\to x$.

\subsubsection*{Coupled Autoregressive Models}
Coupled autoregressive models are an extension of basic autoregressive models, intended to represent the dynamics of systems where multiple time series influence each other. In these models, the value of a variable at a given time point is not only a function of its own previous values but also depends on the past values of other variables in the system. Here, we study the following system consisting of two bidirectionally coupled autoregressive processes of the first order:
\begin{align}
    x(t+1) &= 0.5 x(t) + 0.2 y(t) + \epsilon_x(t), \nonumber \\
    y(t+1) &= Cx(t) + 0.7 y(t) + \epsilon_y(t),
\end{align}
where $C$ is the strength of the coupling between $x$ and $y$, and $\epsilon_x(t)$ and $\epsilon_y(t)$ are drawn from a normal (Gaussian) distribution with a mean of 0 and $\sigma_x^2=\sigma_y^2=0.1$.  Higher values of $C$ represent stronger couplings from $x\to y$, and for $C=0$, the system is unidirectional (only the past of $y$ has an impact on the future of $x$). We examined values of $C\in [0,0.6]$ and created sets of bivariate time series of length $L=300,000$ for each value of $C$ (after a burn-in time of $30,000$ points). The initial conditions of the system were generated from the normal distribution with zero mean and unit variance. 

\begin{figure*} 
  \centering
  \includegraphics[width=13cm]{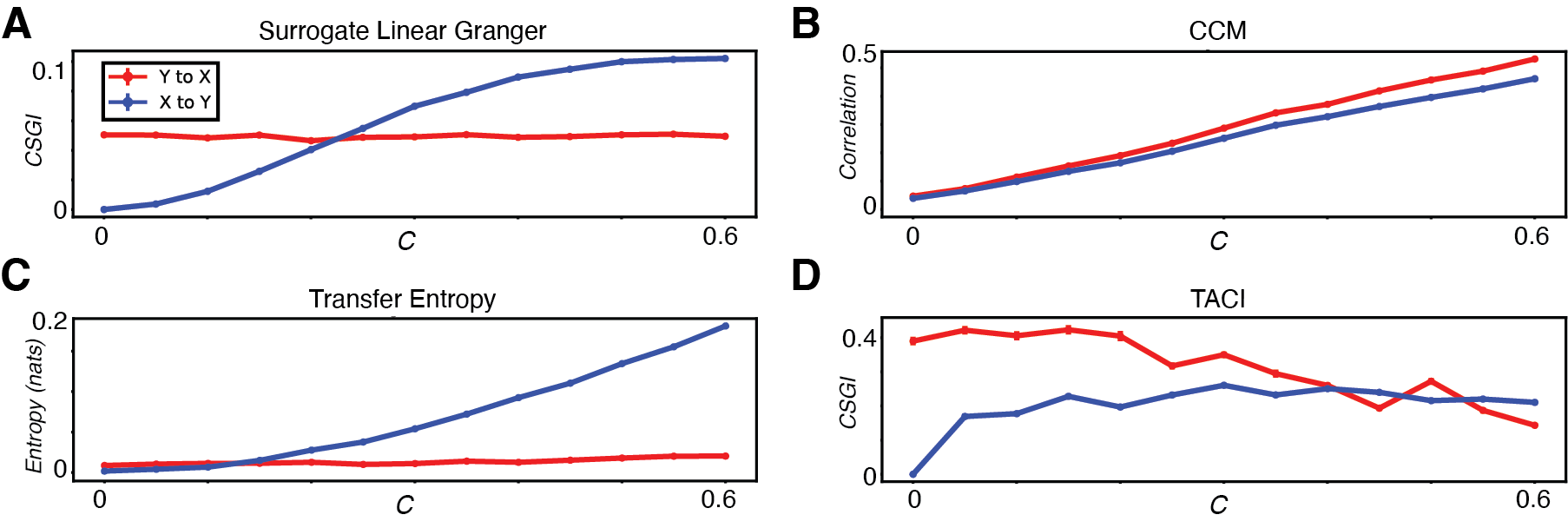}
  \caption[Causal inference in the coupled autoregressive models system]{Causal inference in the coupled autoregressive models system. \textbf{A-D}) Results from applying the four methods to the coupled autoregressive models system.  Error bars are generated using a bootstrapping procedure (see Materials and Methods).}
  \label{fig:ar-processes}
\end{figure*}

Fig. \ref{fig:ar-processes} shows that SLGC does very well at identifying the onset of bi-directionality for $C>0$, with the coupling of $x\to y$ monotonically increasing with $C$. This fact is perhaps not surprising, as SLGC is based on precisely such linear systems.  TACI also does a comparable job at detecting bi-directionality, even roughly predicting the switchover between $x\to y$ and $y\to x$ coupling strengths at $C=0.5$.  CCM, however, does not predict any coupling from $y\to x$ at $C=0$, and TE does not predict any significant coupling from $y\to x$ across all values of $C$.
 
\subsubsection*{Coupled H\'enon Maps}
Our last stationary example, the H\'enon  map, is a well-known example of a discrete-time dynamical system that exhibits chaotic behavior that was first developed as a simplified version of the Poincaré map of the Lorenz model \cite{Henon_original}, and in its chaotic regime, it is characterized by an attractor with a warped horseshoe shape. Here we consider a case of two H\'enon maps, $\vec{x}$ and $\vec{y}$, with unidirectional coupling \cite{bivariate_eqs_examples}:
\begin{align} \label{eqn:henon}
   x_1(t+1) &= 1.4 - x_1^2(t) + 0.3 x_2(t), \\  \nonumber
   x_2(t+1) &= x_1(t), \\  \nonumber
   y_1(t+1) &= 1.4 - \left(Cx_1(t)y_1(t) + (1-C)y_1^2(t)\right) \\ \nonumber
   &\hspace{.2in}+0.3y_2(t), \\  \nonumber
   y_2(t+1) &= y_1(t) \nonumber,
\end{align}
where $C$ controls the strength of the coupling from $\vec{x}$ to $\vec{y}$. For coupling strengths above $C>0.65$, the systems start to show evidence of intermittent synchronizations. This on-off behavior becomes a fully synchronized state after $C>0.7$ \cite{henon_synchronization}.  For $C\in [0,0.9]$, we generated sequences of length 300,000 (after a burn-in period of 30,000) and analyzed data from $x_1$ and $y_1$ for each of the methods.
\begin{figure*} 
  \centering
  \includegraphics[width=13cm]{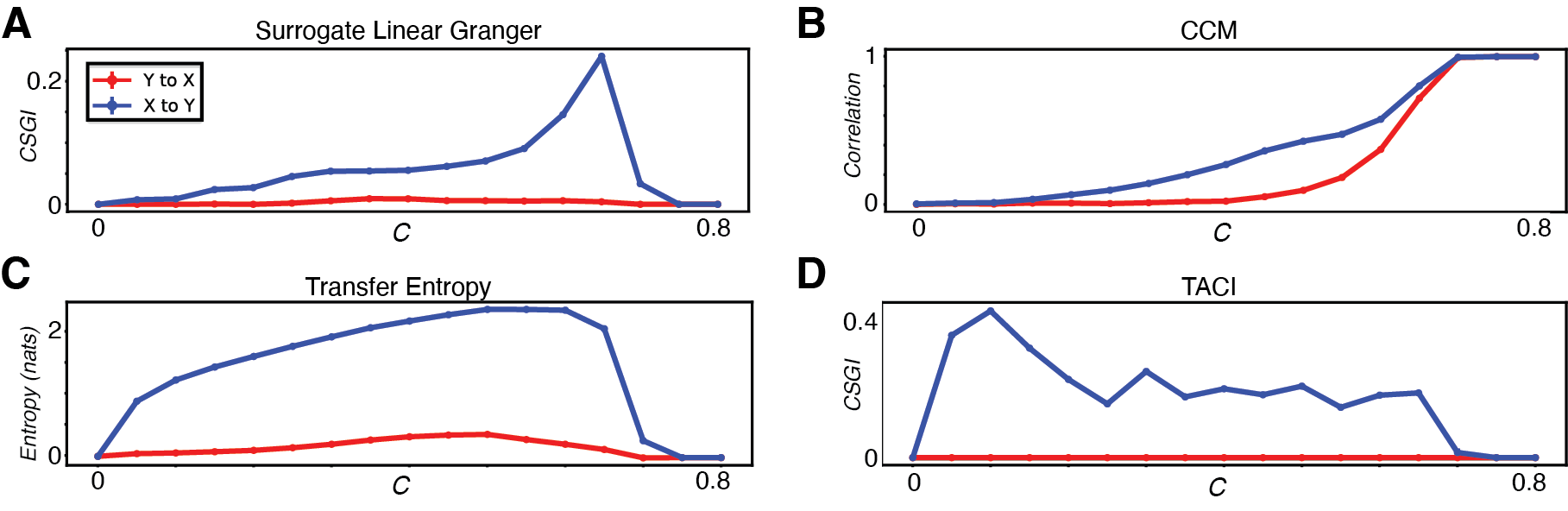}
  \caption[Causal inference in the coupled H\'enon Maps system]{Causal inference in the coupled H\'enon Maps system. \textbf{A-D}) Results from applying the four methods to the coupled H\'enon Maps system. Here, only TACI accurately predicts univariate coupling across all values of $C$ prior to synchronization. Error bars are generated using a bootstrapping procedure (see Materials and Methods).}
  \label{fig:henon-map}
\end{figure*}
TACI is the only method out of the four that correctly identifies the uni-directional coupling between from $\vec{x}$ to $\vec{y}$ (but not from $\vec{y}$ to $\vec{x}$), although SLGC is very close, it is statistically significantly different from zero at intermediate values of $C$. TE and CCM both predict bi-directional interactions, albeit with weaker coupling from $\vec{y}$ to $\vec{x}$ than in the reverse direction.

\subsubsection*{Non-stationary Coupled H\'enon Maps}
TACI is the only method that performed well across all four artificial test cases, but the challenge remains as to whether it can identify patterns in data that change over time.  To test this idea, we generated time series from the coupled H\'enon maps in (\ref{eqn:henon}) but with time-varying couplings, $C_{xy}(t)$ and $C_{yx}(t)$:  
\begin{equation} \label{eqn:henon-time}
\begin{aligned} 
{x_1(t+1)} &= 1.4-C_{yx}(t)y_1(t)x_1(t) \\ 
&\hspace{.2in}+ (1-C_{yx}(t))x_1^2(t) + 0.3x_2\left(t\right),\\ 
{x_2(t+1)} &= x_1\left(t\right),\\ 
{y_1(t+1)} &= 1.4-C_{xy}(t)x_1\left(t\right)y_1\left(t\right) \\
&\hspace{.2in}+ \left(1-C_{xy}(t)\right)y_1^2\left(t\right) + 0.3y_2\left(t\right),\\ 
{y_2(t+1)} &= y_1\left(t\right).\\ 
\end{aligned}
\end{equation}
Here, the two coupling terms are similar to the coupling term, $C$, in (\ref{eqn:henon}) but with time-varying values and potentially allowing for coupling from $y$ to $x$.  

We performed four different tests to see how TACI performs when causal interactions alter with time: \emph{(i)} setting $C_{yx}(t)=0$ and toggling $C_{xy}(t)$ between 0 and 0.6, \emph{(ii)} initially setting $C_{yx}(t)=0$ and $C_{xy}(t)=0.6$ and then switching the two half-way through the run, \emph{(iii)} setting $C_{yx}(t)=0$ and toggling $C_{xy}(t)$ between 0 and 0.6 but with pulses of $C_{xy}(t)=0.6$ being set to different time widths, and \emph{(iv)} setting $C_{yx}(t)=0$ and stepping $C_{xy}(t)$ from 0 to 0.4 and back down again in steps of 0.1. 

Other than the coupling changes, all time series were generated in an identical manner to the previous section.  It is important to note that the network for TACI was only trained once on the entire time series, not specifically for each testing window. Thus, by creating a robust model, our network is able to identify complex causal dynamics that change in time without having to constantly fit new models, as would be the case for SLGC, CCM, and TE.

\begin{figure*}[htbp]
  \centering
  \includegraphics[width=11cm]{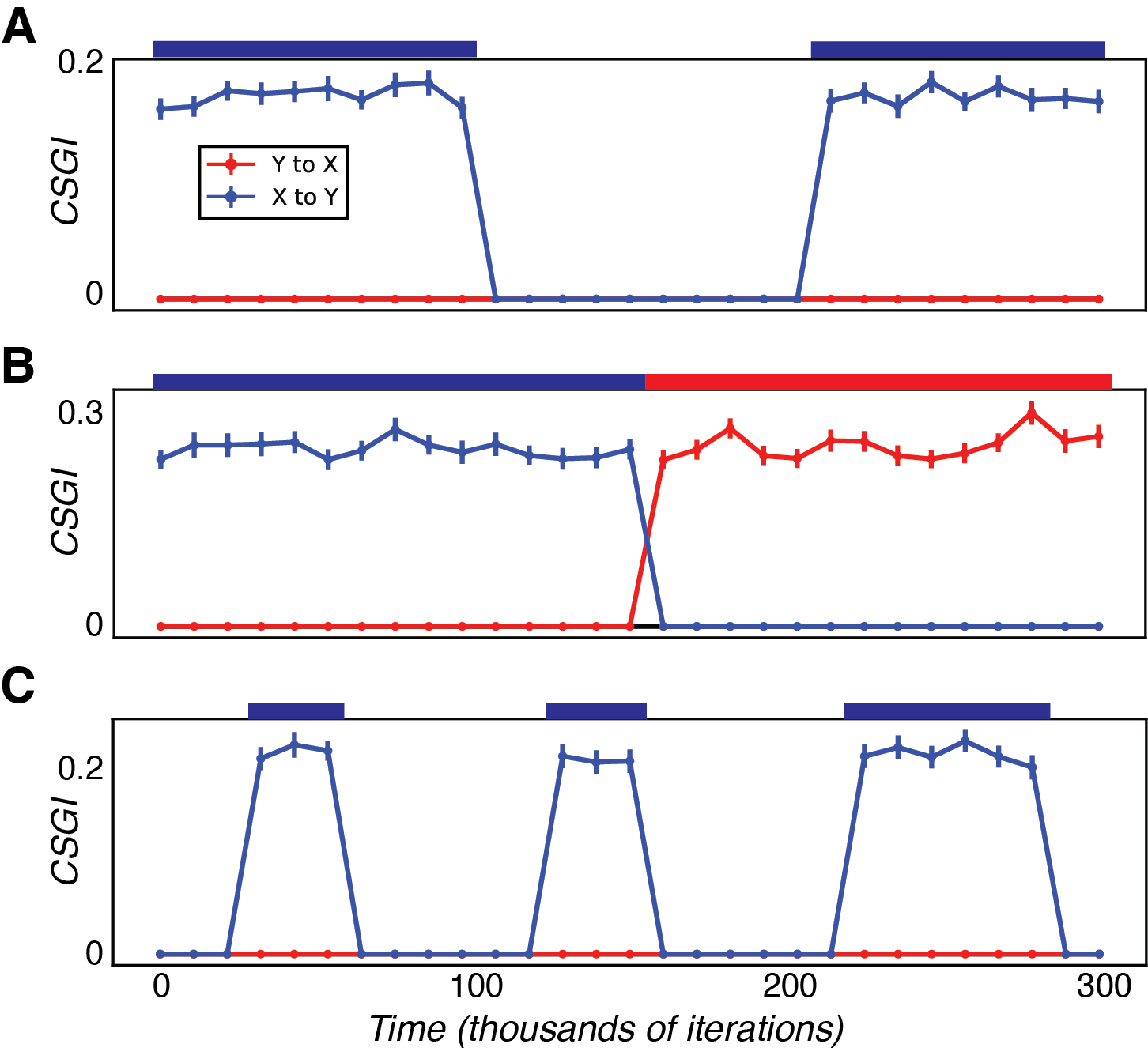}
  \caption[TACI applied to coupled non-stationary H\'enon Maps]{TACI applied to coupled non-stationary H\'enon Maps.  \textbf{A}) A plot of the TACI inference when applied to the coupled H\'enon Maps system where the coupling from $X\to Y$ is set to either $C_{xy}=0.6$ (blue bar above the plot) or $C_{xy}=0$ (no bar above the plot).  \textbf{B}) Same as \textbf{A} but with a toggle from $C_{xy}=0.6$ to $C_{yx}=0.6$ (where the blue and red bars above the plot flip). \textbf{C}) Same as \textbf{A} but with multiple pulses of $C_{xy}=0.6$ of varying sizes.  Error bars are generated using a bootstrapping procedure (see Materials and Methods).}
  \label{fig:henon-non-stationary}
\end{figure*}

\begin{figure*}[htbp]
  \centering
  \includegraphics[width=13cm]{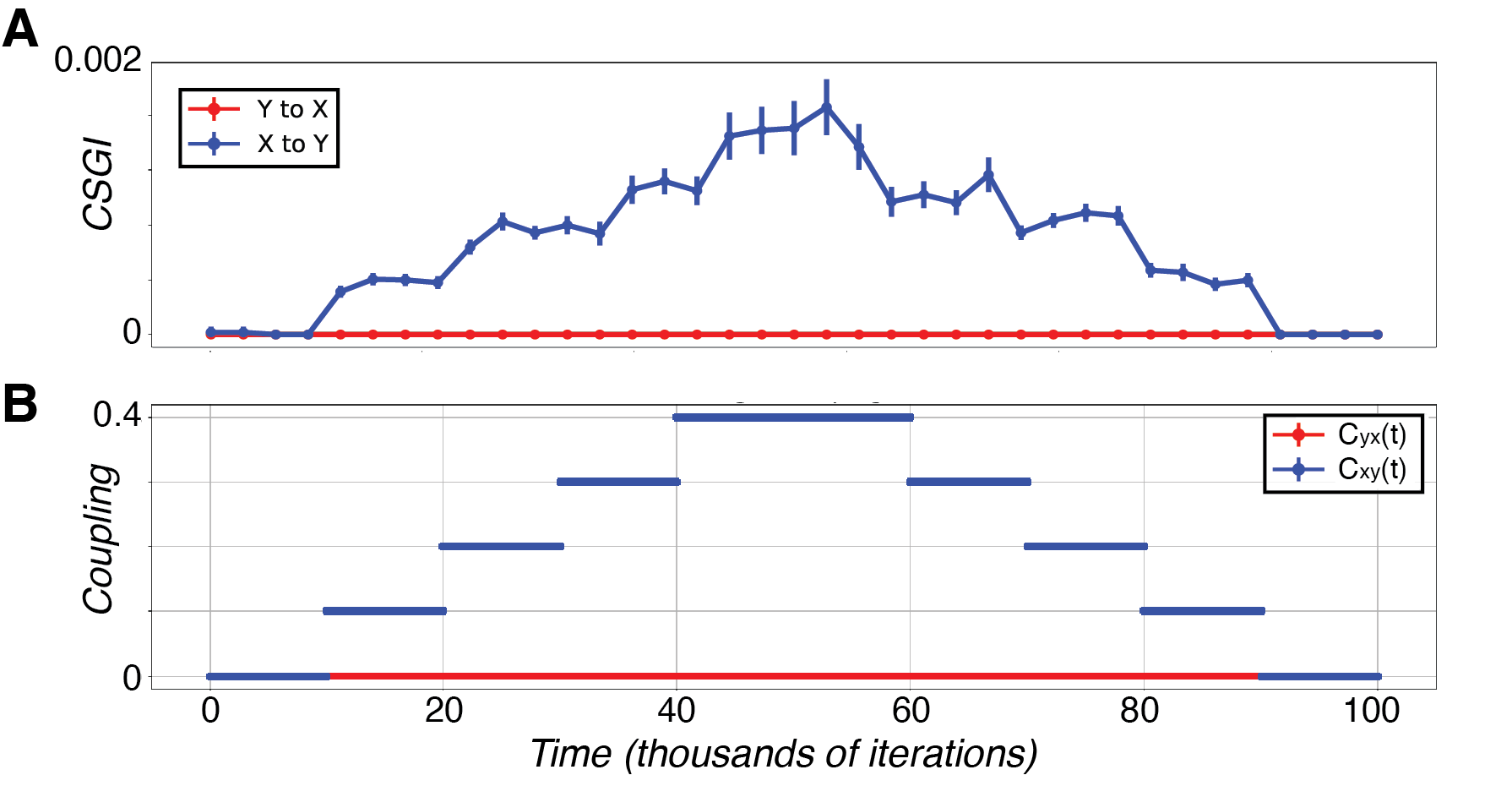}
  \caption[TACI applied to coupled non-stationary H\'enon Maps with ramped couplings]{TACI applied to coupled non-stationary H\'enon Maps with ramped couplings. \textbf{A}) Inferred causal coupling as a function of time during the simulation. \textbf{B} Time series of how the coupling from $X$ to $Y$ was stepped up and then down.  Error bars are generated using a bootstrapping procedure (see Materials and Methods).}
  \label{fig:changing-coupling}
\end{figure*}

In Fig. \ref{fig:henon-non-stationary}, we show that TACI performs well in the first three of these scenarios, ably identifying when eliminations of causal interactions occur, as well as when $C_{yx}(t)=0$ and $C_{xy}(t)$ flip. In addition, Fig. \ref{fig:changing-coupling} shows that the TACI network is able to identify how coupling strengths change with time.

\paragraph{Summary of Results on Artificial Test Systems} 

Among the methods tested, only TACI is able to robustly infer known causal interactions between variables without incorrectly predicting non-existent interactions. TACI consistently differentiates between unidirectional and bidirectional coupling in low, moderate, and strong settings. Additionally, it accurately detects instances when the time series become synchronized in all tested scenarios. TACI excels in identifying complex causal dynamics that evolve over time, such as those observed in pulse systems with time-varying coupling. Given these successes in artificial systems, we will now apply the method to two real-world examples.

\subsection*{Jena Climate Dataset}
\begin{figure*}[htbp]
  \centering
  \includegraphics[width=13cm]{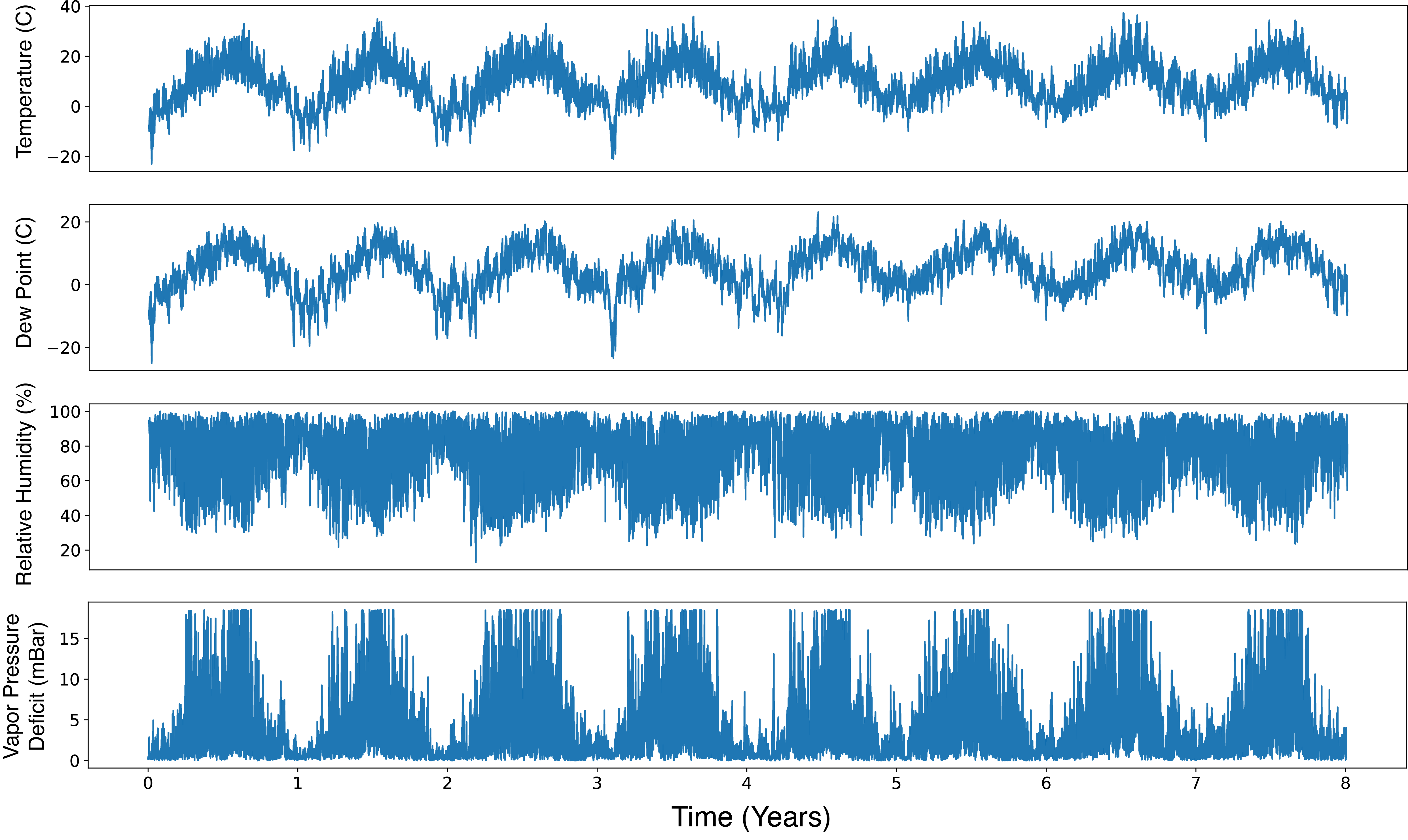}
  \caption{Time series of Temperature, Dew Point, Relative Humidity, and Vapor Pressure Deficit from the Jena Climate Dataset.}
  \label{fig:jena_time_series}
\end{figure*}

The first data set we will test our model on is the ``Jena Climate Dataset", a detailed collection of weather measurements recorded by the Max Planck Institute for Biogeochemistry from a weather station located in Jena, Germany \cite{jena}. The dataset spans nearly eight years -- from January 10, 2009, to December 31, 2016 -- and includes 14 distinct meteorological features recorded every 10 minutes. These features include a wide range of atmospheric conditions, from temperature to relative humidity to vapor pressure deficit (see Table \ref{tab:jena_climate_features} for details).  Several example time series are shown in Fig. \ref{fig:jena_time_series}.
\begin{table}[!htbp]
    \centering
    \caption{Summary of Jena Climate Dataset Features}
    \begin{tabular}{lc}
        \toprule
         Feature                & Description \\
        \midrule
         Date/Time       & Date-time reference \\
         p (mbar)                & Atmospheric pressure stated millibars \\
         T (degC)                  & Temperature in Celsius \\
         Tpot (K)                   & Temperature in Kelvin \\
         $T_{dew}$ (C)                & Dew Point Temperature in Celsius \\
        $R_H$ (\%)                    & Relative Humidity in percentage \\
         VPmax (mbar)                & Saturation vapor pressure \\
         VPact (mbar)                 & Vapor pressure \\
         VPdef (mbar)                & Vapor pressure deficit \\
         sh (g/kg)                 & Specific humidity \\
        H$_2$O C (mmol/mol)             & Water vapor concentration \\
         rho (g/m\(^3\))              & Air density \\
         wv (m/s)                  & Wind speed \\
         max. wv (m/s)                & Maximum wind speed \\
         wd (deg)                     & Wind direction in degrees \\
        \bottomrule
    \end{tabular}
    \label{tab:jena_climate_features}
\end{table}

\begin{figure*}[htbp]
  \centering
  \includegraphics[width=13cm]{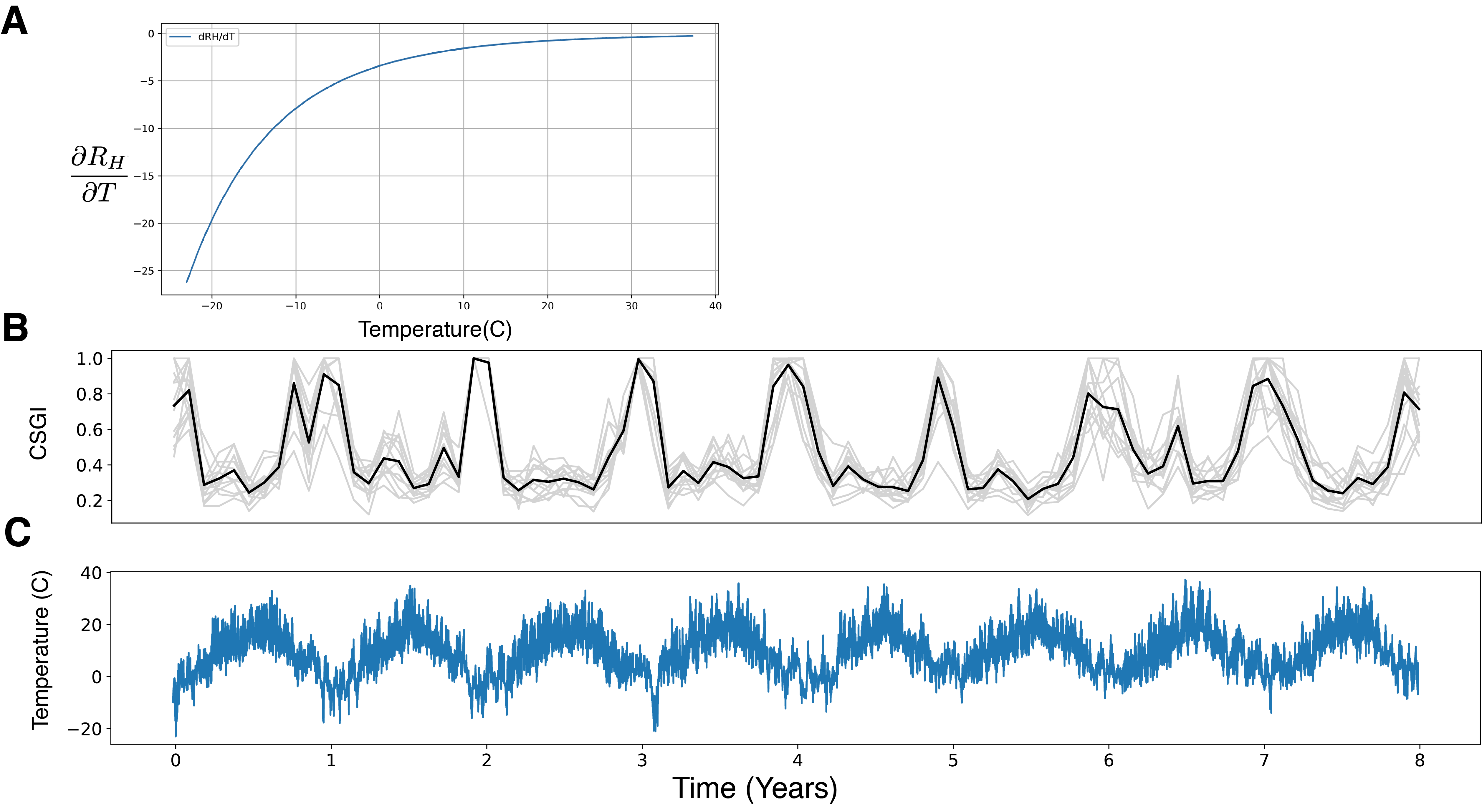}
  \caption[Causal interactions with relative humidity from the Jena Climate Dataset]{Causal interactions with relative humidity from the Jena Climate Dataset. \textbf{A}) Empirical relationship between relative humidity and air temperature (assumes $T_{dew}=10$).  Note the large negative partial derivative at low values of $T$. \textbf{B}) TACI predictions for causal interactions for how the other 13 variables in Table \ref{tab:jena_climate_features} affect relative humidity as a function of time across the eight years of the dataset (gray lines, mean trajectory is the black line).  Note how causal influence peaks consistently when the temperature (\textbf{C}) is at its nadir, just as predicted by the plot in \textbf{A}.}
  \label{fig:jena-causality}
\end{figure*}

A key advantage of these data is that some of the interactions are known already due to empirical models of atmospheric dynamics, providing a good test case for our method on real data.  One example is the relationship between relative humidity ($R_H$), the dew point ($T_{dew}$), and the temperature ($T$), which is given by
\begin{equation}
    R_H = 100 \exp\frac{17.625 T_{dew}}{T_{dew} + 243.04} \Bigg/ \exp\frac{17.625  T}{T + 243.04},
\end{equation}
where $T_{dew}$ and $T$ are in degrees Celsius and $R_H$ is a percentage \cite{RH_equation}. Calculating the partial derivative of $R_H$ with respect to $T$ (keeping $T_{dew}$ fixed), we find that we should expect stronger interactions to occur from $T$ to $R_H$ at lower temperatures (Fig. \ref{fig:jena-causality}A). After training our TACI model from each of the variables in the data set onto $T$, we indeed find that causal interactions peak during epochs when the temperature drops (Fig. \ref{fig:jena-causality}B), showing that our method can accurately find temporal variations in causal interactions in messy real-world data.

\subsection*{Electrocorticography in Non-Human Primates}
Lastly, we used electrocorticography (ECoG) data from non-human primates to test whether our methodology can detect time-varying interactions between brain regions from these electrophysiological signals.  These data exhibit extraordinarily complex dynamics that shift in time as an animal changes its state: from sleep to wake, from satiated to hungry, from attending from one object to another, and so on \cite{buzsaki2006rhythms}.  These alterations are often subtle, and, thus, understanding how different regions of the brain drive one another's activity requires a method that can detect how slight variations in the relationship between variables lead to changing interactions across time. 

Here, we analyzed publicly available ECoG data from a single monkey (\emph{Macaca fuscata}) \cite{Granger_monkey_data,CCM_monkey_data,Nagasaka_monkey_data}.  These recordings consisted of 128 channels of data that recorded activity from a hemisphere of the monkey's brain that covered the visual, temporal, parietal, motor, prefrontal, and somatosensory cortices, sampling at 1kHz (details can be found in \cite{Granger_monkey_data}).  Data were collected during both awake and anesthetized states to examine neural activity across different consciousness levels.  To generate an anesthetized state, the monkey was chair-restrained and propofol was injected intravenously. The recording sessions were structured into four distinct phases: an initial phase where the monkey is awake with eyes open, a subsequent phase where the monkey is awake but with its eyes covered, a phase where the monkey is under deep anesthesia, induced to reach a state of loss of consciousness, and a final stage where the monkey recovers from anesthesia with its eyes covered. The depth of anesthesia was assessed by monitoring the monkey's responsiveness to tactile stimulation and the presence of slow wave oscillations in the ECoG signal \cite{Granger_monkey_data}.  

Previous studies analyzing these data for changes in causal interactions using Spectral Granger Causality \cite{Granger_monkey_data} or CCM \cite{CCM_monkey_data}, but each was only able to analyze data at the level of the four phases described in the previous paragraph (each requiring training a separate model, as well).  Specifically, we trained TACI on one monkey (George in \cite{Nagasaka_monkey_data}) with a sequence length of 50 to account for the extended autocorrelation time observed in the time series (average of 53). Approximately 53 minutes of data corresponding to the four previously outlined phases were utilized for this purpose. The training was conducted over 300 epochs or until the point of convergence. Further details of the parameters used can be found in \ref{table:parameters}  

Finally, to compare with these previous studies, while we calculated the causal interactions between each pair of electrodes, we will present many of the results as the average result between pairs of electrodes assigned to the same region of the cortex.  Here, we will be using the eight coarse-grained regions defined in \cite{CCM_monkey_data}: the medial prefrontal cortex (mPFC), lateral prefrontal cortex (lPFC), pre-motor cortex (PMC), motor and somatosensory cortex (MSC), temporal cortex (TC), parietal cortex (PC), higher visual cortex (HVC), and lower visual cortex (LVC).

\begin{figure*}
  \centering
  \includegraphics[width=13cm]{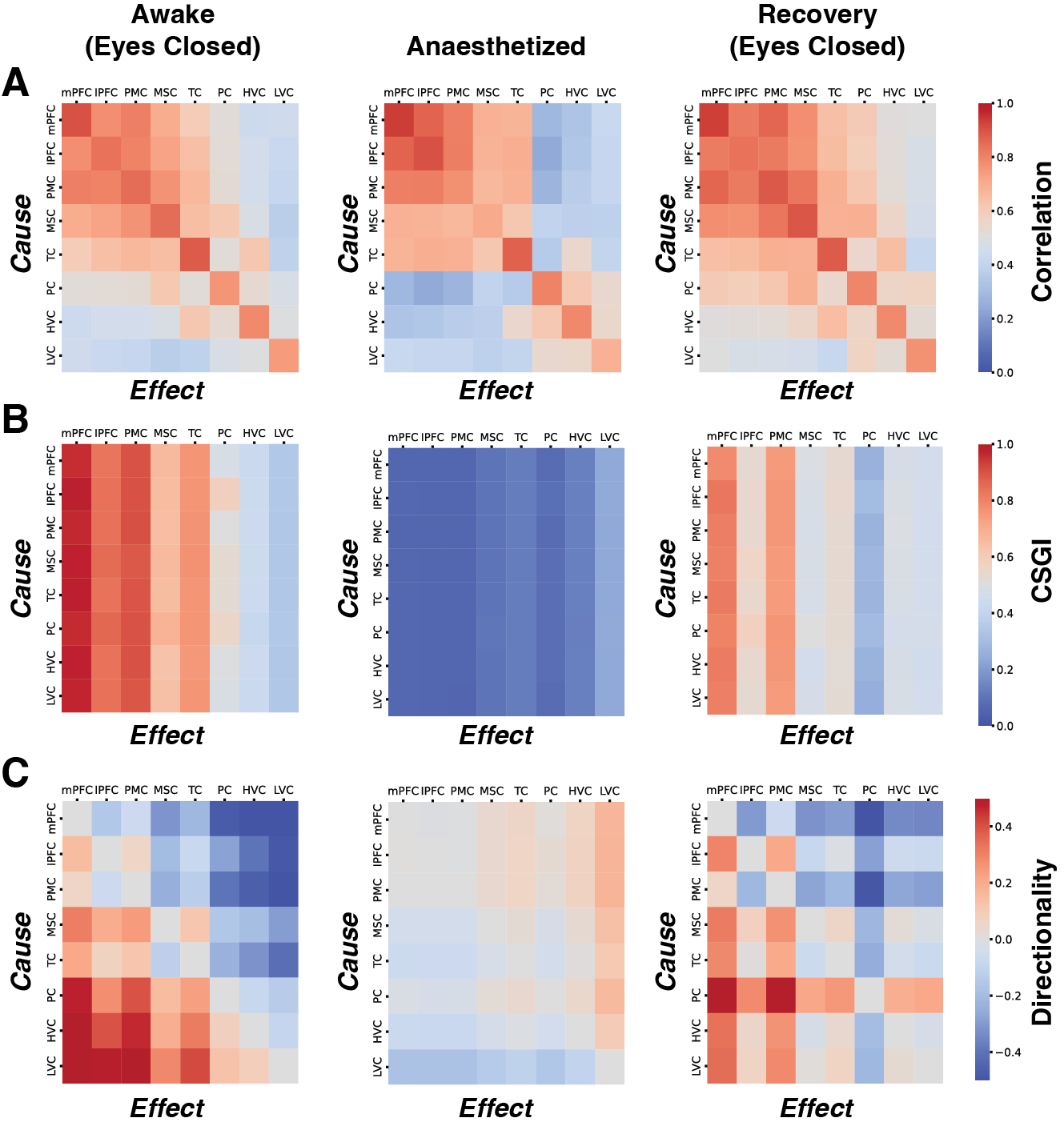}
  \caption[Interactions between brain regions in ECoG data]{Interactions between brain regions in ECoG data. Each plot here shows the average interaction between all electrodes within each of the 8 coarse-grained regions described in the text.  The left matrices are from before the anesthesia was administered, the middle matrices are from when the monkey was anesthetized, and the right plots are from the recovery period. \textbf{A} is the Pearson correlation between the signals, \textbf{B} is the TACI-derived inference of causal interaction, and \textbf{C} displays the TACI Directionality -- the difference between the CSGI score in one direction minus the CSGI score in the other direction.}
  \label{fig:monkey-coarse-causality}
\end{figure*}

Fig. \ref{fig:monkey-coarse-causality} shows time-averaged values of correlation (\textbf{A}), TACI-derived causal interactions (\textbf{B}), and Directionality (\textbf{C}), which we define as the difference in CSGI values in one direction vs. the other, for epochs of time before, during, and after anaesthetization.  For correlation, we measure the average Pearson correlation coefficient between all the electrodes assigned to the various regions.  Note that the diagonal terms do not necessarily have to be equal to one here, as electrodes within a region are not perfectly correlated with one another. There are only minimal changes in brain region interactions across the three time windows when measuring correlation, but large differences emerge when analyzing the data using TACI. Specifically, we see that almost all interactions disappear during the anesthetized period, with the interactions beginning to re-emerge during the recovery period. These results differ from the results from CCM in \cite{CCM_monkey_data}, where they claimed that while some interactions decreased, others strengthened (this effect is seen in our Directionality measurements, however). Also interesting are the nearly vertical lines in Fig. \ref{fig:monkey-coarse-causality}B, implying that certain regions like the mPFC might be affected broadly by signals from various parts of the cortex -- a finding that agrees with the commonly held notion that the mPFC's role often involves higher-level cognitive function \cite{Anastasiades.2021}.  Again, it should be noted that only one TACI network was trained per pair of interactions across all time epochs, unlike the other methods we describe, which must find interactions separately during each measurement period.

\begin{figure*}
  \centering
  \includegraphics[width=13cm]{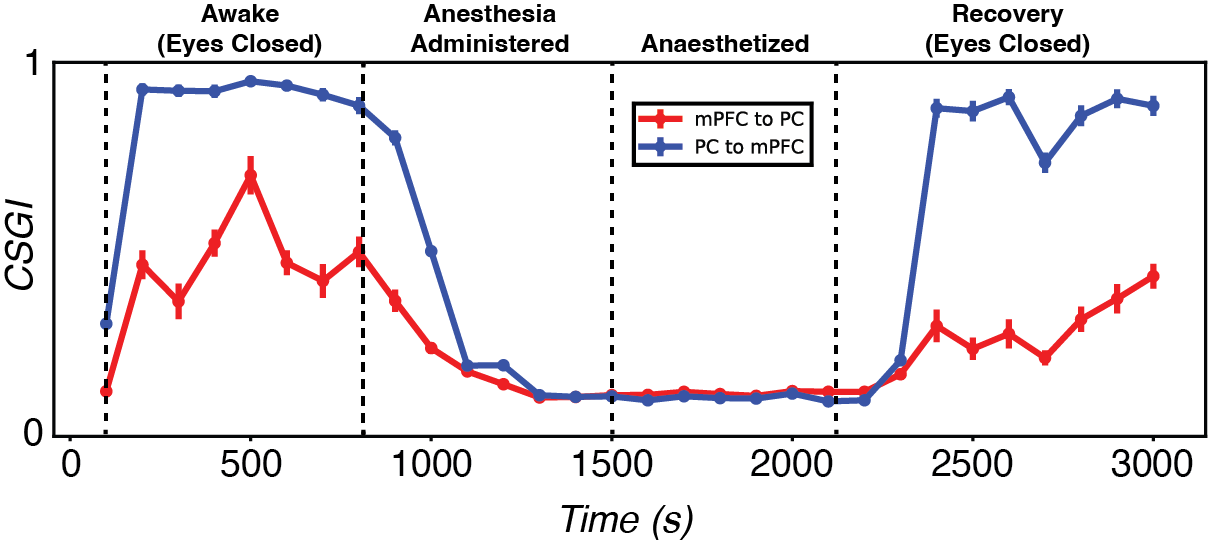}
  \caption[Causal interactions across time between Parietal and medial Prefrontal Cortices]{Causal interactions across time between Parietal and medial Prefrontal Cortices. Plot of the average TACI-derived interactions between PC and mPFC over the course of the anesthesia experiment.  Error bars are the standard errors of the mean across all electrode interactions, and the dashed lines represent change points in the experimental protocol (labeled above the axes).}
  \label{fig:monkey-channels}
\end{figure*}

Lastly, taking advantage of the aforementioned property of TACI, we took a finer-grained look at how interactions between a pair of regions might change with time during the experiment, specifically the mPFC and the PC.  In Fig. \ref{fig:monkey-channels}, we show how these regions' interactions alter with time.  Using our approach, we observe how the coupling slowly decays upon administration of the propofol and how it rapidly increases a few minutes into the recovery period.  Also interesting is that while during the awake periods, PC consistently has a casual interaction towards mPFC, the reverse interaction has significant temporal fluctuations whose study might lead to insights into how these brain regions drive each other during cognitive tasks.

\section*{Discussion}
In this article, we introduce a new methodology for probing time-varying causal interactions in complex dynamical systems using a novel machine learning architecture for causal inference, Temporal Autoencoders for Causal Inference (TACI), combined with a novel metric for assessing causal interactions using surrogate data.  A particular advantage of our approach is being able to train a single model that captures the dynamics of the time series across all points in time, allowing for time-varying interactions to be found without retraining, a computationally expensive endeavor for most artificial neural networks.  We found that our method performed well compared to other methods in the field on synthetic data sets with known causal interactions, including those with time-varying couplings between variables.  We also found that our method was able to identify known interactions between variables in a climate data set and was able to discover subtle temporal fluctuations in coupling in non-human primate ECoG data.

Our approach, while novel, is not without its limitations. One of the primary concerns is the extensive training time and the resource-intensive nature of the model. Implementing TACI, especially on large datasets, requires significant computational power and time.  We envision that several technical improvements in the network architecture and training will allow for the method to be sped-up considerably, however.  Another concern is the potential for overfitting due to TACI's considerable modeling capacity. While the framework is designed to capture the nuanced dynamics of causal relationships over time, like most other causal network models, this method can fit data too closely if not trained properly, resulting in models that perform exceptionally well on training data but generalize poorly to unseen data.  Furthermore, TACI incorporates elements of the Granger causality approach, which means it also inherits some of its problems. Granger causality assumes that the causal variable contains unique information about the future values of the effect variable, which might not always hold true in complex systems where numerous latent factors influence outcomes.  Lastly, but importantly, as our approach is based solely on observational data, TACI only attempts to provide hypotheses about causal relationships between variables or to infer important relationships between variables when perturbation experiments are impossible or unethical to perform.

These limitations withstanding, however, the results presented in this chapter provide evidence that our approach will be broadly applicable to complicated data sets with time-varying causal structure, with particular promise for neural data, where we hope to build our understanding of how parts of the brain shift their interactions as behavioral states and needs alter in the world.

%\section*{Conclusion}

%For more information, see \nameref{S1_Appendix}.

\section*{Materials and methods}

At its core, TACI uses a two-headed autoencoder architecture implemented in a two-step process aiming to facilitate the prediction of future states and the inference of causal relationships between different time series datasets. In the first application, the two-headed autoencoder is utilized to process the original time series data, $x(t)$ and $y(t)$. The encoder segments of this autoencoder independently process $x(t)$ and $y(t)$, capturing and encoding their temporal dynamics and features into latent representations. These representations are then merged in the bottleneck, combining the distilled information from both time series into a unified latent space that encapsulates potential causal interactions. From this combined latent representation, the decoder works to reconstruct or predict the future trajectory of $y(t)$, shifted by a time $\tau$. The second application involves replacing $x(t)$ with the surrogate data $x^{(s)}(t)$. This surrogate data is generated to mimic the statistical properties of $x(t)$ but is designed to break any potential causal link between $x(t)$ and $y(t)$

This two-step process is essential for figuring out how these variables are linked to one another. The model can validate the presence of a causal relationship by comparing the predictive accuracy of the decoder when using the original $x(t)$  versus the surrogate $x^{(s)}(t)$. A significant drop in accuracy with the surrogate data suggests that the original $x(t)$ contains specific information causally linked to the future states of $y(t)$. 

\subsection*{Architecture}

In the TACI architecture, the concept of a two-headed encoder is employed to simultaneously process two distinct time series datasets, denoted as $x(t)$ and $y(t)$. This design allows for the independent yet parallel analysis of each time series, enabling the model to capture and encode their individual characteristics and temporal dynamics before merging their representations during the bottleneck process. The input sequences are selected to be greater in length than the autocorrelation time of each variable. This ensures that the sequences capture meaningful temporal dependencies and dynamics. A GaussianNoise layer is added to enhance the model's ability to generalize and prevent overfitting. 

The most important part of the encoder includes the use of a Temporal Convolutional Network (TCN) layer.  Thus,  capturing the long-term dependencies within each time series. This layer utilizes several key parameters: \textit{``nb\_filters"} sets the number of convolutional filters, \textit{``kernel\_size"} affects the temporal extent of each convolution, \textit{``dilations"} allows the model to efficiently gather information across various temporal distances. Additionally, \textit{``Dropout"} layers are used to decrease overfitting by randomly dropping units during the training phase. Following the TCN, a \textit{Conv1D} layer continues to process the data for each series, allowing the network to change dimensionality while preserving temporal resolution. An \textit{AveragePooling1D} layer may then downsample the \textit{Conv1D} layer's output by pooling across the temporal dimension. This operation reduces the sequence length, emphasizing significant features and further decreasing data dimensionality. The data is subsequently processed by a series of \textit{Dense} layers that compress it into a dense, lower-dimensional latent representation. The size of these layers decreases in each successive layer, concentrating the information into a more compact form. 

The bottleneck stage starts once the two-headed encoder has finished processing and compressing the input sequences into a lower-dimensional latent space representation. The Bottleneck merges these latent representations through an element-wise multiplication operation. By combining the representations in this manner, the model effectively captures the potential interactions and dependencies between the time series, which are essential for uncovering causal relationships.

Once the latent representations are merged in the Bottleneck, this combined representation is forwarded to the Decoder. The Decoder's task is to predict the future trajectory of the target time series. The first step in the Decoder is to progressively upscale the combined latent representation. This is achieved through a series of \textit{Dense} layers, where each layer aims to increase the dimensionality of the data. The number and size of these layers are determined by the complexity of the data and the level of compression achieved by the Encoder. After the initial upscaling, an \textit{UpSampling1D} layer is used to increase the sequence length to its original size, effectively reversing the pooling operation performed in the Encoder. A TCN layer is used to ensure that the reconstructed data maintains its temporal integrity and dynamics. This layer mirrors the TCN configuration in the Encoder, utilizing the same parameters for \textit{``nb\_filters"}, \textit{``kernel\_size"}, and \textit{``dilations"} to capture the temporal dependencies and patterns necessary for accurate prediction. Lastly, a Dense output layer produces the final prediction of the future states of the target time series. 

\subsection*{Training and Prediction}

As discussed earlier, the training phase of the TACI model involves four distinct configurations of the network. Central to this phase is the use of the Mean Squared Error (MSE) as the loss function, which facilitates the optimization of predictions for future trajectories against actual observed values. The Adam optimizer \cite{kingma2014adam} is employed for its adaptive learning rate capabilities. Training is performed across 300 epochs to give the model enough time for the parameters to adjust and converge toward optimal solutions. The parameters controlling the batch size and data shuffling are finely tuned to balance computational efficiency and the promotion of model generalization. Callbacks such as ReduceLROnPlateau, EarlyStopping, and ModelCheckpoint are employed in this phase for optimizing the training process by adjusting learning rates, preventing overfitting, and preserving the best model state, respectively.

Surrogate data were created by drawing random values from a uniform distribution between zero and one until a time series the length of the original one was generated.  An alternative method would have been to create a surrogate time series by first converting the original series into the frequency domain through a Fourier transform. Then, we could apply random phase shifts,  making sure the amplitude spectrum remained unaltered. This randomness is crucial to breaking any specific temporal dependencies present in the original series. Following this process, an inverse Fourier transform could be employed to reconstruct the series back into the time domain. This step generates a new time series that mirrors the original in terms of its overall power distribution but only has random contingencies with its partner data set.  In practice, however, we found that this latter methodology did not result in more accurate results in training TACI, so we focused on the initially described method for generating surrogate data in this study.

After training is completed, the model moves on to the prediction phase, where the focus shifts to evaluating the trained model.   In the first step of the prediction phase, the pre-trained models are loaded, each representing a unique configuration designed in the training phase to capture and analyze the causal dynamics between the time series datasets $x(t)$ and $y(t)$. At the same time, the full original dataset is divided into sequences with the same length and structure as the models were trained on.   The prediction process occurs over defined rolling windows to allow for a temporal exploration of the dataset, enabling the models to make predictions for future states of the time series within each window.   The models' accuracy in forecasting future time series states is quantitatively evaluated for each rolling window using the $R^2$ metric.   To enhance the reliability and confidence of these assessments, 100 bootstrap samples are generated for each window.   The causal inference for each rolling window can be determined using the CSGI Eq. \ref{CSGI}. Through this calculation, the model not only quantifies the strength and direction of the causal relationship but also shows its variation over time, providing a dynamic and temporal perspective on causal inference.  

For each interval, a bootstrap strategy is implemented. This strategy involves creating a set number of surrogate samples by randomly resampling within the interval. These samples are then used to evaluate the model's predictions, which are generated under two conditions: one using the actual interactions between the time series and another using the surrogate data. By employing Equation \ref{CSGI}, it's possible to derive scores from which we compute both the mean and standard deviation. These computations provide insight into the average performance and variability of the model's predictions across the bootstrap samples. The utilization of bootstrap methods significantly enhances the analytical depth by ensuring that the derived error bars and confidence intervals are supported by a solid statistical foundation. These statistics play a vital role in establishing the error bars in the plotted figures. By repeating this procedure across all intervals, the method provides a comprehensive view of how model performance fluctuates over time and under different conditions.

% \subsection{Limitations}

% \section*{Supporting information}

% % Include only the SI item label in the paragraph heading. Use the \nameref{label} command to cite SI items in the text.
% \paragraph*{S1 Fig.}
% \label{S1_Fig}
% {\bf Describe S1 Figure 1.} Descriptive text (optional).

%\subsection*{TACI Network Parameters}

\begin{table*}
\centering
\begin{tabular}{l p{8cm} l}
\hline
\textbf{Parameter} & \textbf{Description} & \textbf{Value} \\ \hline
nb\_filters & Number of filters in TCN layers. & 32 \\
kernel\_size & Size of the kernel in TCN layers. & 32 \\
dilations & Dilation rates for TCN layers. & [1, 2, 4, ..., 32] \\
nb\_stacks & Number of stacked TCN layers. & 1 \\
ts\_dimension & Dimensionality of the time series. & 1 \\
dropout\_rate\_tcn & Dropout rate for TCN layers. & [0.0, \dots, 0.5] \\
dropout\_rate\_hidden & Dropout rate for hidden layers. & [0.0, \dots, 0.5] \\
conv\_kernel\_init & Kernel initializer for convolutional layers. & 'he\_normal' \\
latent\_sample\_rate & Downsampling rate in the latent space. & 2 \\
act\_funct & Activation function used in layers. & 'elu' \\
epochs & Number of training epochs. & 300 \\
batch\_size & Batch size for training. & 512 \\
shuffle & Whether to shuffle the data during training. & [True or False] \\
scaling\_method & Method used for scaling the input data. & Z score \\
loss\_funct & Loss function used for training. & 'mse' \\
noise & Standard deviation of Gaussian noise added. & [0.0, \dots, 0.5] \\
window\_len & Size of the rolling window for predictions. & value \\
seq\_length & Length of sequences used for training/prediction. & [10, \dots , 100] \\
lag & Lag between \(x(t)\) and \(y(t)\) for prediction. & [10, \dots , 100] \\ \hline
\end{tabular}
\caption[Parameters used in the TACI model training and prediction phases]{Parameters used in the TACI model training and prediction phases (ranges indicate the parameter range used across the examples in this chapter)}
\label{table:parameters}
\end{table*}

% \subsubsection*{Coupled Bi-directional Two Species Model}

\section*{Acknowledgments}
Both authors were supported by the Human Frontier Science Program (RGY0076/2018) and the Simons Foundation (707102 \& 876207), and JC was supported by the NSF Physics of Living Systems Student Research Network (PHY-1806833).  GJB would like to acknowledge the Aspen Center for Physics, where many of the initial ideas for this work were generated.

%\nolinenumbers

% Either type in your references using
% \begin{thebibliography}{}
% \bibitem{}
% Text
% \end{thebibliography}
%
% or
%
% Compile your BiBTeX database using our plos2015.bst
% style file and paste the contents of your .bbl file
% here. See http://journals.plos.org/plosone/s/latex for 
% step-by-step instructions.
% 
%\begin{thebibliography}{10}

%\bibitem{bib1}
%Conant GC, Wolfe KH.
%\newblock {{T}urning a hobby into a job: how duplicated genes find new
%  functions}.
%\newblock Nat Rev Genet. 2008 Dec;9(12):938--950.

%\bibitem{bib2}
%Ohno S.
%\newblock Evolution by gene duplication.
%\newblock London: George Alien \& Unwin Ltd. Berlin, Heidelberg and New York:
%  Springer-Verlag.; 1970.%

%\bibitem{bib3}
%Magwire MM, Bayer F, Webster CL, Cao C, Jiggins FM.
%\newblock {{S}uccessive increases in the resistance of {D}rosophila to viral
%  infection through a transposon insertion followed by a {D}uplication}.
%\newblock PLoS Genet. 2011 Oct;7(10):e1002337.

%\end{thebibliography}

\clearpage
%\bibliography{references}  

\end{document}